\documentclass[twoside,noinfoline,11pt]{article}

\usepackage{jmlr2e}
\usepackage{bm}
\usepackage{graphicx}

\jmlrheading{1}{2020}{1-8}{01/2020}{}{}{Jos\'{e} A. Ferreira}

\firstpageno{1} 

\begin{document}

\title{Models under which random forests perform badly; consequences for applications}

\author{\name José A. Ferreira \email jose.ferreira@rivm.nl \\
       \addr Department of Statistics, Informatics and Modelling\\
                National Institute for Public Health and the Environment (RIVM)\\
                Antonie van Leeuwenhoeklaan 9,
                3721 MA Bilthoven, The Netherlands}

\editor{}

\maketitle

%\vspace{-0.15cm}
\begin{abstract}%   <- trailing '%' for backward compatibility of .sty file
We give examples of data-generating models under which Breiman's random forest may be extremely slow to converge to the optimal predictor or even fail to be consistent. The evidence provided for these properties is based on mostly intuitive arguments, similar to those used earlier with simpler examples, and on numerical experiments. 
Although one can always choose models under which random forests perform very badly, we show that simple methods based on statistics of `variable use' and `variable importance' can often be used to construct a much better predictor based on a `many-armed' random forest obtained by forcing initial splits on variables which the default version of the algorithm tends to ignore.
\end{abstract}


\section{Introduction}

Breiman’s random forest \citep{Breiman:2001} is a feasible and flexible algorithm for constructing nonparametric statistical predictors. Nowadays it is acknowledged to be easy to use and to perform very well in general, even in problems involving many predictor variables (see the survey by  \cite{BiauScornet:2016} or the introduction to \cite{ScornetBiauVert:2015})---so well, indeed, that several authors have posed and studied the question of its consistency (see \cite{BiauDevroyeLugosi:2008}, \cite{Wager:2014}, \cite{ScornetBiauVert:2015}, and the references provided by these authors).
`Universally consistent' nonparametric statistical predictors
% such as those based on histogram, nearest-neighbour and kernel prediction rules
have been known for a long time (\cite{Nadaraya:1964}, \cite{Watson:1964}, \cite{Stone:1977}, \cite{DevroyeWagner:1980}, \cite{DevroyeGyorfiLugosi:1996}, \cite{GyorfiEtAL:2002}), but their computer implementations tend to be slow, especially when dealing with many variables
(e.g.~\cite{DevroyeGyorfiLugosi:1996}, p.~62). In view of their accuracy and of the high speed of their implementations, random forests would become even more attractive if they were shown to be consistent under general data-generating mechanisms. In particular, consistency, in addition to accuracy, is indispensable in applications of statistical prediction to the estimation of ‘causal effects’ based on observational data (pp.~120-1, 167-9 and subsection 4.1 of \cite{Ferreira:2015}). The simplest and clearest general consistency result on Breiman’s random forest seems to be theorem 1 of \cite{ScornetBiauVert:2015}, which in essence states that if the `response' follows an additive regression model
(e.g.~a linear regression model)
then random forests are consistent in mean-square if the number of terminal nodes (‘leaves’) of the constituent trees increases to infinity at a slower rate than the size of the subsamples on which they are grown and if those subsamples, rather than being bootstrap samples, are drawn without replacement from the full sample.\footnote{\cite{DurouxScornet:2018} show that the number of terminal nodes and the size of the subsamples can have a substantial effect on the finite-sample performance of random forests, and that if the size of the subsamples is about $1-e^{-1}$ of that of the full sample then random forests with trees based on subsamples perform very similarly to Breiman’s random forests with trees based on bootstrap samples. \cite{Wager:2014} presents results of wider scope concerning other variants of random forests under conditions which to us seem more restrictive or more difficult to verify.}This is encouraging, but the counterexample in proposition 8 of \cite{BiauDevroyeLugosi:2008} shows that one particular version of random forest is not consistent in complete generality. Moreover, the arguments around figure 5 of \cite{KimLoh:2001} and figure 1 of \cite{BiauDevroyeLugosi:2008} and those in subsection 2.2 of \cite{ZhuZengKosorok:2015} show that random forests can perform badly, and suggest that they may be inconsistent in some of their versions, under certain so-called `checkerboard-type' models.\\
\indent
The main purpose of this note is to exhibit two classes of models for which random forests
may be extremely slow to converge to the optimal predictor or even be inconsistent. One such class generalizes the checkerboard-type just mentioned; the other is much more general. In both classes, if the response variable $Y$ is a nonconstant function of each of $X_1,X_2,\ldots,X_d$ ($d>2$) but is independent of each of $X_1$ and $X_2$, then Breiman's random forest  typically fails to use both variables to the full even when $(X_1,X_2)$ is the strongest predictor of $Y$. This is a consequence of the ‘one-dimensional greed’ of random forest: each split in a tree is based on the variable that `best explains'  $Y$ out of a randomly drawn subset of predictor variables; but if $Y$ is independent of each of $X_1$ and $X_2$ and repeated conditioning on some of $X_3,\ldots,X_d$ continues to provide information on $Y$ (typically the case when $(X_3,\ldots,X_d)$ has a positive density on an open rectangle) then the algorithm will tend to split at a variable other than $X_1$ and $X_2$ and most trees tend to make little use of $(X_1,X_2)$; cf.~subsection 2.2 of \cite{ZhuZengKosorok:2015}. Although this argument applies only to random forests with tree splits done on one among two or more randomly selected variables, i.e.~to those in which the parameter usually referred to as {\tt mtry} (e.g.~in the {\tt R} package {\tt randomForest} of \cite{LiawWiener:2002}) is $\ge 2$, the algorithm typically requires many splits before $X_1$ and $X_2$ get to be used together in a tree, which can only hinder convergence, even if convergence is to the optimal predictor.\\
\indent
The argument alone is not sufficient to prove the lack of consistency of random forests even in the case where ${\tt mtry}\ge 2$, because if a forest is sufficiently large and the trees in it grow sufficiently tall---say until each of their terminal nodes contains fewer than a fixed number of distinct observations―then $X_1$ and $X_2$ are likely to be picked at a `late' stage during the construction of some of the trees, and one cannot deny that even very late splits on those two variables may compensate for their having been ignored earlier. In fact, it is easy to see (and must be well known, the observation having been used earlier by \cite{IshwaranKogalur:2010}) that if $X_1,X_2,\ldots,X_d$ are all binary then a tree predictor grown until each of its terminal nodes contains only observations with the same data on $(X_1,X_2,\ldots,X_d)$ (which implies that at certain splits all the predictor variables must be tried) is consistent, and so is a random forest of trees grown in that way; and the same is true, of course, if each of
$X_1,X_2,\ldots,X_d$ takes a finite number of values. Accordingly, our examples are not meant for situations in which $(X_1,X_2,\ldots,X_d)$ has finite range, because as soon as the observations in a node all have the same data on $(X_3,\ldots,X_d)$ the next two splits along that node will involve $X_1$ and $X_2$ and eventually the partition corresponding to the resulting tree will be equivalent to a partition which would have started with splits on $X_1$ and $X_2$. It is evident, moreover, that if the $X_j$s have finite range then the partitions generated by such trees are asymptotically equivalent to those obtained by splitting the data without recourse to a criterion involving data on the $(X_j,Y)$s―i.e.~they amount to partitions with the so-called `X-property', and the corresponding trees amount to `partitioning estimates' for which more general consistency results are available (section 20.1 of \cite{DevroyeGyorfiLugosi:1996}, section 4.1 of \cite{GyorfiEtAL:2002}).

Rather, the examples given in section 2 are meant for situations in which the random forest algorithm is properly greedy in one-dimension, i.e.~uses data on $(X_j,Y)$ for some $j$ to create a split; they do not apply to the variant proposed and shown to be consistent in section 20.14 of \cite{DevroyeGyorfiLugosi:1996} and in section 6 of \cite{BiauDevroyeLugosi:2008}, whose rules for splitting are based on data on $(X_1,X_2,\ldots,X_d,Y)$. The arguments we are able to provide in favour of a very slow rate of convergence are intuitive, but they are easily seen to be supported by simple numerical experiments, as illustrated in section 3.

%The arguments we are able to provide in favour of no consistency or of very slow rate of convergence, presented in the appendix, are intuitive; but we have a rigorous, if simple, result for random forests whose trees have a fixed number of nodes. On the other hand, the arguments are supported by simple numerical experiments, as illustrated in section 3.

Although the properties we identify imply that random forests can perform very badly compared to the optimal predictor, it is not at all our intention to put random forests in a bad light (which would be difficult to do in view of their good record in applications and of the scarcity of methods of comparable scope and success). In fact, our second purpose is to show that simple methods based on statistics of `variable importance' and `variable use' can help to determine whether the bad performance of a random forest is due to the presence of predictor variables such as the $X_1$ and $X_2$ just mentioned and to construct a `many-armed' version of random forest that performs much better; this too is explained in section 2 and illustrated by simulation in section 3. Section 4 offers some perspective on our results and considers open questions.

\section{The examples; consequences for applications}

The following is a textbook example, often attributed to S.N. Bernstein (e.g.~\cite{Burrill:1972}, p.~241), of three dependent random variables that are pairwise independent: Let $B_1$, $B_2$ and $B_3$ be independent Bernoulli random variables of parameter $1/2$ and set
\[
X_0:=\mathbf{1} _{\{B_1=B_2\}},\quad X_1:=\mathbf{1} _{\{B_1=B_3\}},\quad X_2:=\mathbf{1} _{\{B_2=B_3\}}.
\]
Then $X_0$ is a function of $X_1$ and $X_2$, namely $X_0=\mathbf{1} _{\{X_1=X_2\}}=\delta_{X_1,X_2}$,
%\[
%X_0=\mathbf{1} _{\{X_1=X_2\}}=\delta_{X_1,X_2},
%\]
\[
\mathbb{P}(X_j=x_j,X_k=x_k )=
\textstyle\frac{1}{4}
%\frac{1}{4}
=\mathbb{P}(X_j=x_j)\mathbb{P}(X_k=x_k ) \quad ( j\neq k),
\]
%\vspace{-0.5cm}
but
\[
\mathbb{P}(X_0=x_0,X_1=x_1,X_2=x_2 )\neq\mathbb{P}(X_0=x_0)\mathbb{P}(X_1=x_1)\mathbb{P}(X_2=x_2)=\textstyle\frac{1}{8}
\]  
for $x_0,x_1,x_2\in\{0,1\}$, so $X_0$, $X_1$ and $X_2$ are pairwise independent but not independent.

Now put
\begin{equation}\label{equation1}
Y:=\delta_{X_1,X_2} f(X_3,\ldots,X_d,\epsilon)+(1-\delta_{X_1,X_2})g(X_3,\ldots,X_d,\zeta)+h(X_3,\ldots,X_d,\eta),
\end{equation}
where $\epsilon$, $\zeta$, $\eta$ are random variables and $(X_3,\ldots,X_d)$ is a random vector, all four independent and also independent of $(X_0,X_1,X_2)$, and $f$, $g$ and $h$ are real-valued functions.
Writing $\mathbf{X}=(X_1,X_2,\mathbf{X}' )$, $\mathbf{X}'=(X_3,\ldots,X_d)$, and, for real numbers $x_1,x_2,\ldots,x_d$, $\mathbf{x}=(x_1,x_2,\mathbf{x}')$, $\mathbf{x}'=(x_3,\ldots,x_d)$, we have
%\begin{eqnarray*}\label{equation2}
%\mathbb{P}(Y\le y|X=x)&\!\!\!\!=\!\!\!\!&
%\mathbb{P}\left(\delta_{x_1,x_2}f(\mathbf{x}',\epsilon)+(1-\delta_{x_1,x_2})g(\mathbf{x}',\zeta)+
%h(\mathbf{x}',\eta)\le y\left|{X_1=x_1,X_2=x_2}\atop{\mathbf{X}'=\mathbf{x}'}\right.\right)\\
%&\!\!\!\!=\!\!\!\!&\delta_{x_1,x_2}\mathbb{P}(f(\mathbf{x}',\epsilon)+h(\mathbf{x}')\le y)+
%(1-\delta_{x_1,x_2})\mathbb{P}(g(\mathbf{x}',\zeta)+h(\mathbf{x}',\eta)\le y).
%\end{eqnarray*}
\begin{equation}\label{equation2}
\mathbb{P}(Y\le y|\mathbf{X}=\mathbf{x})=
\delta_{x_1,x_2}\mathbb{P}(f(\mathbf{x}',\epsilon)+h(\mathbf{x}',\eta)\le y)+
(1-\delta_{x_1,x_2})\mathbb{P}(g(\mathbf{x}',\zeta)+h(\mathbf{x}',\eta)\le y).
\end{equation}
Evidently, the best predictor of $Y$ based on all the variables except $\epsilon$, $\zeta$ and $\eta$ is provided by the function $\mathbf{x}\rightarrow\mathbb{P}(Y\le y|\mathbf{X}=\mathbf{x})$, e.g.~in the form of $\mathbb{E}(Y|\mathbf{X}=\mathbf{x})$ or med$(Y|\mathbf{X}=\mathbf{x})$ when $Y$ is numeric proper.
On the other hand, for $j=1,2$
\[
\mathbb{P}(Y\le y|X_j=x_j,\mathbf{X}' =\mathbf{x}')=
\frac{1}{2} \mathbb{P}(f(\mathbf{x}',\epsilon)+h(\mathbf{x}',\eta)\le y)+\frac{1}{2}\mathbb{P}(g(\mathbf{x}',\zeta)+h(\mathbf{x}',\eta)\le y)
\]
by the independence of $X_0$ and $X_j$. Similarly, for $j=1,2$
\[
\mathbb{P}(Y\le y|X_j=x_j )=
\frac{1}{2}\mathbb{P}(f(\mathbf{X}',\epsilon)+h(\mathbf{X}',\eta)\le y)+\frac{1}{2} \mathbb{P}(g(\mathbf{X}',\zeta)+h(\mathbf{X}',\eta)\le y)
\]
by the independence of $X_0$, $X_j$ and $\mathbf{X}'$.
In particular, $Y$ is independent of $X_1$, and independent of it also conditionally on $\mathbf{X}'$; and likewise $Y$ is independent of $X_2$, and independent of it also conditionally on $\mathbf{X}'$. Since in general
$\mathbb{P}(Y\le y|X_j=x_j)$ and $\mathbb{P}(Y\le y|X_j=x_j,\mathbf{X}'=\mathbf{x}')$ provide predictors of $Y$ that are worse than those provided by
$\mathbb{P}(Y\le y|\mathbf{X}=\mathbf{x})$, which are optimal, any predictor that misses out on one of $X_1$ and $X_2$ will be suboptimal.

Now each split of each tree involved in a random forest is determined by selecting, among a random subset of {\tt mtry} predictors, the variable that `best explains' $Y$, unconditionally or conditionally on some of the predictors. If ${\tt mtry}\ge 2$ then, since $Y$ is independent of each of $X_1$ and $X_2$ separately, unconditionally as well as conditionally on some of $X_3,\ldots,X_d$, unless one of $X_1$ and $X_2$ has been selected at an earlier split the random forest algorithm will tend to select none of them again but instead one of $X_3,\ldots,X_d$. Consequently, a large proportion of the trees grown by the algorithm should involve only predictor variables among $X_3,\ldots,X_d$ in a large proportion of their terminal nodes, and the resulting forest is worse than a `two-armed' forest of trees grown upon a first split based on $(X_1,X_2)$. When ${\tt mtry}=1$ it is more likely that one of $X_1$ and $X_2$ be selected at a split, but if $d$ is `'large' then the probability that both $X_1$ and $X_2$ be involved in a terminal node must be $<1$ even for very large $n$, so even for very large $n$ there should be a non-negligible proportion of the trees that involve only predictor variables among $X_3,\ldots,X_d$ in a non-negligible proportion of their terminal nodes, and again the resulting forest should perform worse than a two-armed forest and hence worse than the optimum predictor.

%As suggested earlier, this is no more than an intuitive argument indicating that random forests can be inconsistent or very slow to converge to the best predictor.

Our second class of models features the same type of relations between $Y$, $(X_1,X_2)$ and $(X_3,\ldots,X_d)$ but is much more general. Consider a random vector $(X_0,X_1,X_2)$ with probability density function
\begin{equation}\label{equation3A}
f(x_0,x_1,x_2)=f_0 (x_0) f_1 (x_1) f_2 (x_2 )\left\{1-\varphi(x_0,x_1,x_2)\right\}
\end{equation}
for densities $f_0$, $f_1$ and $f_2$ and some function $\varphi$ not identically equal to zero and such that $\varphi\le 1$ and
$\int_{\mathbb{R}}f_0 (x_0 ) f_1 (x_1 ) f_2 (x_2) \varphi(x_0,x_1,x_2)\mbox{d}x_j=0$
for all $j$, so that $g_{j,k} (x_j,x_k):=\int_{\mathbb{R}} f(x_0,x_1,x_2)\mbox{d}x_l=f_j (x_j ) f_k (x_k )$ for $j,k,l$ all different and again $X_0$, $X_1$ and $X_2$ are pairwise independent without being independent.\footnote{This example must be well known, but we do not recall a textbook where we may have seen it before.} For instance, one may take $X_0$, $X_1$ and $X_2$ symmetric ($f_0$, $f_1$ and $f_2$ even) and
% $\varphi(x_0,x_1,x_2)=\sin x_0 \sin x_1 \sin x_2$ or
$\varphi(x_0,x_1,x_2)=\prod_{j=0}^2 x_j (1+x_j^2)^{-1/2}$;
% $\varphi(x_0,x_1,x_2)=x_0 x_1 x_2/\sqrt{(1+x_0^2)(1+x_1^2)(1+x_2^2)}$;
%\[
%\varphi(x_0,x_1,x_2)=\sin x_0 \sin x_1 \sin x_2;
%\]
by transforming the $X_j$s by ${\tilde X}_j=T_j (X_j)$, say, one obtains ${\tilde X}_0$, ${\tilde X}_1$ and ${\tilde X}_2$ with arbitrary marginal distributions that remain pairwise independent without being independent. Notice that in principle it is easy to simulate $(X_0,X_1,X_2)$ by the `rejection method'.

Assuming that $X_0$, $X_1$ and $X_2$ dependent but pairwise independent have been defined, one can then set
$\mathbf{X}^{''}\equiv (X_1^{''},X_2^{''}):=(X_1,X_2)$ and
\begin{equation}\label{equation3B}
X_0^{''}=H^{-1}(\xi;\mathbf{X}^{''}),
\end{equation}
where $\xi$ is a standard uniform variable independent of all the other variables mentioned so far and
$H(x;x_1,x_2 )=\mathbb{P}(X_0\le x|\mathbf{X}^{''}=(x_1,x_2 ))$,
$H^{-1}(u;x_1,x_2 )=\min\left\{x:H(x;x_1,x_2 )\ge u\right\}$
($0<u<1$), to get a vector $(X_0^{''},\mathbf{X}^{''})\equiv(X_0^{''},X_1^{''},X_2^{''})$ with the same joint distribution as $(X_0,X_1,X_2)$ in which the first coordinate is a function of the other two and of $\xi$. Finally,
\begin{equation}\label{equation3C}
Y:=\Psi(H^{-1} (\xi;\mathbf{X}^{''}),\mathbf{X}',\varepsilon),
\end{equation}
where $\mathbf{X}'$ is independent of $\mathbf{X}^{''}$, $\varepsilon$ is independent of all the other variables, and $\Psi$ is some function, defines a model in which $Y$ is dependent on $\mathbf{X}^{''}$ but independent of each of its coordinates, and for which a random forest predictor based on data on
$(\mathbf{X}',\mathbf{X}^{''})$ may be very slow to converge to the optimum and perhaps even be inconsistent.

Evidently, this example may be generalized to more than three variables (e.g.~by adding several independent versions of the right-hand side of (\ref{equation3C})), leading to a model in which the response is dependent on a finite set of random variables but independent of each of them.

Finally, we can generalize (\ref{equation3A})  by replacing each of $X_1$ and $X_2$ above by random vectors
${\mathbf X}_1$ and ${\mathbf X}_2$ taking values in ${\mathbb R}^{d_1}$ and ${\mathbb R}^{d_2}$ such that $X_0$, ${\mathbf X}_1$ and ${\mathbf X}_2$ are dependent but pairwise independent, which yields a $Y$ dependent on the corresponding
$\mathbf{X}^{''}=(\mathbf{X}_1^{''},\mathbf{X}_2^{''})$ but independent of each of 
$\mathbf{X}_1^{''}$ and $\mathbf{X}_2^{''}$. In this case, for a tree-based algorithm to approximate the optimal predictor it must pick `sufficiently many' variables in $\mathbf{X}_1^{''}$ and $\mathbf{X}_2^{''}$ during the construction of its trees, and that may be quite difficult if $\max\{d_1,d_2\}$ is large and
the independent covariates $\mathbf{X}^{'}$ provide some information on $Y$. 

For a simple, concrete example let $\phi_d$ denote the standard normal density on ${\mathbb R}^d$ and $\mathbf{1}_d$ a $d$-vector of 1s, and let $X_0$, ${\mathbf X}_1$ and ${\mathbf X}_2$ have density
\[
f(x_0,\mathbf{x}_1,\mathbf{x}_2)=\phi_1(x_0)\phi_{d_1}(\mathbf{x}_1)\phi_{d_2}(\mathbf{x}_2)
\left(1-\frac{x_0}{\sqrt{c_0^2+x_0^2}}\frac{\mathbf{x}_1\cdot \mathbf{1}_{d_1}}{\sqrt{c_1^2+(\mathbf{x}_1\cdot \mathbf{1}_{d_1})^2}}
\frac{\mathbf{x}_2\cdot \mathbf{1}_{d_1}}{\sqrt{c_2^2+(\mathbf{x}_2\cdot \mathbf{1}_{d_2})^2}}\right),
\]
$\mathbf{x}_1\in\mathbb{R}^{d_1}\!$, $\mathbf{x}_2\in\mathbb{R}^{d_2}\!$, $c_0,c_1,c_2$ constants. Then all the marginal distributions are standard normal, 
$X_0$ is independent of ${\mathbf X}_1$ and independent of ${\mathbf X}_2$ but dependent on 
$({\mathbf X}_1,{\mathbf X}_2)$, and we can define $X_0^{''}$ and $Y$ by the corresponding versions of (\ref{equation3B}) and (\ref{equation3C}) with ${\mathbf X}^{''}=({\mathbf X}_1,{\mathbf X}_2)$ and a $d_3$-vector ${\mathbf X}'$ independent of all the other variables. In the resulting model the optimal predictor conditional on ${\mathbf X}_1$ and on ${\mathbf X}'$ is a function of the latter vector alone, and so is the optimal predictor conditional on ${\mathbf X}_2$ and on ${\mathbf X}'$, while the optimal predictor
is a function of $({\mathbf X}_1,{\mathbf X}_2,{\mathbf X}')$.
Thus, depending on the sample size, on $d_1$, $d_2$ and $d_3$, and on how 
$\Psi(\cdot,{\mathbf x}',\cdot)$ varies with ${\mathbf x}'$, an ordinary tree predictor will tend to use
 ${\mathbf X}_1$ and ${\mathbf X}_2$ insufficiently and belatedly; in contrast, `many-armed' tree predictors and associated random forests made up of an initial partition of the range of
${\mathbf X}_1$ and ${\mathbf X}_2$ with trees grown atop will typically perform much better and may have a chance of approaching the optimal predictor.

\indent
\subsection{A possible remedy}
Now it is clear that in a random forest the coordinates of ${\mathbf X}_1$ and ${\mathbf X}_2$ in our last model always have a chance of showing their worth, and the extent to which they are worthy predictors should become apparent in variable importance statistics {\it irrespectively of their frequency of occurrence in the trees}. If ${\mathbf X}_1$ and ${\mathbf X}_2$ include strong predictors then one can check how frequently they are used in the trees by looking at appropriate statistics such as the proportion of trees making use of them, the average number of terminal nodes per tree in which they are involved, etc.; if such {\it variable usage statistics} show that certain predictors are less frequently used than one might expect given their importance, then their use probably needs to be enforced at an earlier stage.\footnote{As far as we know, the implementations of random forests currently available provide no `ready-made' information on variable usage, but some partial information can sometimes be extracted from them, as illustrated in section \ref{NumericalIllustration}. We emphasize that measures of variable importance quantify the improvement in accuracy that results from using the various predictor variables, but they provide no information about how often a variable is used in relation to its importance.}  Expressed with this latitude, the truth of these statements seems evident to us; but, of course, in any given problem the extent to which one will be able to identify vectors like ${\mathbf X}_1$ and ${\mathbf X}_2$ as important predictors worth splitting on beforehand depends on the other elements involved---sample size, $d_1$, $d_2$ and $d_3$, etc. Thus, in a given problem and with a given sample size one may or may not be able to recognize the importance of such variables; but if their importance is found to be substantial then one may as well check the corresponding usage statistics. Evidently, the combined use of variable importance and variable usage statistics is not useful only with data that follow exactly a model of the type considered here;
it is plausible that in real-life problems---for instance in genetics, where haplotypes rather than genotypes can predict phenotypes---there will be sets of variables that are approximately analogous to 
${\mathbf X}_1$ and ${\mathbf X}_2$ and admit of similar remedies.\\
\indent
It is easy to use our examples to simulate data and illustrate how badly random forests can perform compared to the optimal predictor, even with very large sample sizes; the last model especially can present problems to random forests because it undermines what normally is their 
strength, namely the possibility of finding advantageous partitions of the range of the predictor variables by trying univariate splits at a time.\footnote{We do not want to suggest that other algorithms perform better on such data; the more classic Nadaraya-Watson or nearest-neighbour algorithms, for example, will generally not perform better unless $d_1$, $d_2$ and $d_3$ are `small'. A simple {\tt R} script for simulating data from the last model and comparing the performance of random forests on it with that of the optimal predictor, as well as scripts that reproduce the results of section 3, may be obtained from the author.} 
As said earlier, however, the illustrations provided in this paper are not meant to show random forests at their worst; they are mostly based on simulation from a simple version of (\ref{equation1}) which is hardly unfavourable to random forest and yet shows a clear gap between it and the best predictor. 

\section{Numerical illustration}\label{NumericalIllustration}
We consider the following special case of (\ref{equation1}):
\begin{equation}\label{equation8}
Y:=\delta_{X_1,X_2} f(X_3,\ldots,X_d,\epsilon)+\{1-\delta_{X_1,X_2}\}g(X_3,\ldots,X_d,\zeta)+\eta, 
\end{equation}
with $d=10$, $f(X_3,\ldots,X_d,\epsilon)=\alpha_3 X_3+\cdots+\alpha_d X_d+\epsilon$,
$g(X_3,\ldots,X_d,\zeta )=\beta_3 X_3+\cdots+\beta_d X_d+\zeta$, ${\boldsymbol\alpha}:=(\alpha_3,\ldots,\alpha_{10})=(1,2,\ldots,8)/8$, ${\boldsymbol\beta}:=(\beta_3,\ldots,\beta_{10})=3{\boldsymbol\alpha}/4$, $\epsilon$, $\zeta$  and $\eta$  independent standard normal random variables, $(X_1,X_2)$ as in section 2 and independent of $(\epsilon,\zeta,\eta)$, and $(X_3,\ldots,X_{10})$ normally distributed with covariance matrix $\Sigma=(2^{-|j-k|})_{j,k=1,..,10}$, mean vector equal to diag($\Sigma$), and independent of all the other variables.\\
\indent
By the result of \cite{ScornetBiauVert:2015} we might expect random forests constructed with data $(X_1,\ldots,X_{10},Y)$ such that $\delta _{X_1,X_2}=1$ to be consistent for the first branch of the model, namely $f(X_3,\ldots,X_{d},\epsilon)$, and random forests constructed with data such that $\delta _{X_1,X_2}=0$ to be consistent for the second branch,
$g(X_3,\ldots,X_{d},\zeta)$.
Thus we might expect the {\it two-armed random forest predictor} that consists of two random forests, one for predicting the response of new data satisfying $\delta _{X_1,X_2}=1$ and the other for predicting the response of new data satisfying
$\delta _{X_1,X_2}=0$, to converge in mean-square to the right-hand side of (\ref{equation8}) without the error terms as the size $n$ of the data set used to construct it increases to infinity; in particular, we might expect its law given
${\mathbf X}={\mathbf x}\in\mathbb{R}^d$ to converge to the corresponding right-hand side of (\ref{equation2}) as $n\rightarrow\infty$, and the mean-square error of the predictor to approach
$\mbox{Var}(\epsilon+\eta)=\mbox{Var}(\zeta +\eta)=2$. The random forest predictor, on the other hand, may decrease its mean-square error as $n$ increases but we do not expect it to attain the minimal value.\\
\indent
Figure 1 summarizes the results of two simulations, one based on training and test samples of size $n=10,000$, the other on training and test samples of size $n=100,000$; the random forests were constructed with the {\tt R} package {\tt randomForest} of \cite{LiawWiener:2002} using 1000 trees and the default settings. The scatter-plots indicate the agreement between the response of the test set and the corresponding prediction, and the estimates of mean-square error (mse), mean absolute error (mae) and proportion of explained variance quantify the accuracy of the predictions. The top left plot of figure 1 shows the results obtained with the random forest when $n=10,000$; the algorithm, which of course ignores the model generating the data, performs remarkably well. The top right plot ranks the predictor variables according to their relative importance: by our specifications of
${\boldsymbol\alpha}$ and ${\boldsymbol\beta}$ and of the distribution of $(X_3,\ldots,X_{10})$, the rank of $X_j$ for $j\ge 3$ is equal to $j$, and the higher the rank, the higher the importance; the algorithm ranks the variables correctly except, as we shall see below, for $X_1$ and $X_2$.\footnote{We use Breiman's measure of variable importance as implemented in the {\tt randomForest} package with {\tt scale=TRUE}; variable importance is then not really an estimate of the percent worsening of the mean square error that results from a random permutation of the data on a variable but
a scaled version of it. As is well known, variable importance must be regarded as a relative measure which quantifies how much more important each variable is relative to the others.} The bottom panel summarizes the results obtained with the random forest and with the supposedly consistent two-armed random forest when $n=100,000$; while the two-armed random forest appears to have practically attained the optimum, the random forest does not seem to improve upon a mean-square error of about 2.5.\\
\indent
The first panel of figure 2 summarizes the results obtained separately with the two random forests making up the two-armed random forest based on the training data set of size 100,000. Both random forests appear to be close to reaching the optimum value of 2 for the mean-square error when $n=100,000$. Note that the optimal predictor of $Y$ based on $(X_3,\ldots,X_{10})$ in the sense of mean-square error is $\gamma \cdot(X_3,\ldots,X_{10})$ with $\gamma=({\boldsymbol\alpha}+{\boldsymbol\beta})/2$. If the trees in the random forest would make little use of $X_1$ and $X_2$ one might expect the random forest predictor to be close to it. This is confirmed by the bottom panel of figure 2, which summarizes the performance of $\gamma\cdot(X_3,\ldots,X_{10})$ and compares the predictions of the latter with those of the forest.\\

\begin{figure}[hbtp]\label{figure1}
\begin{center}
\includegraphics[clip,trim=0 0 0 0,angle=0,width=1\textwidth]{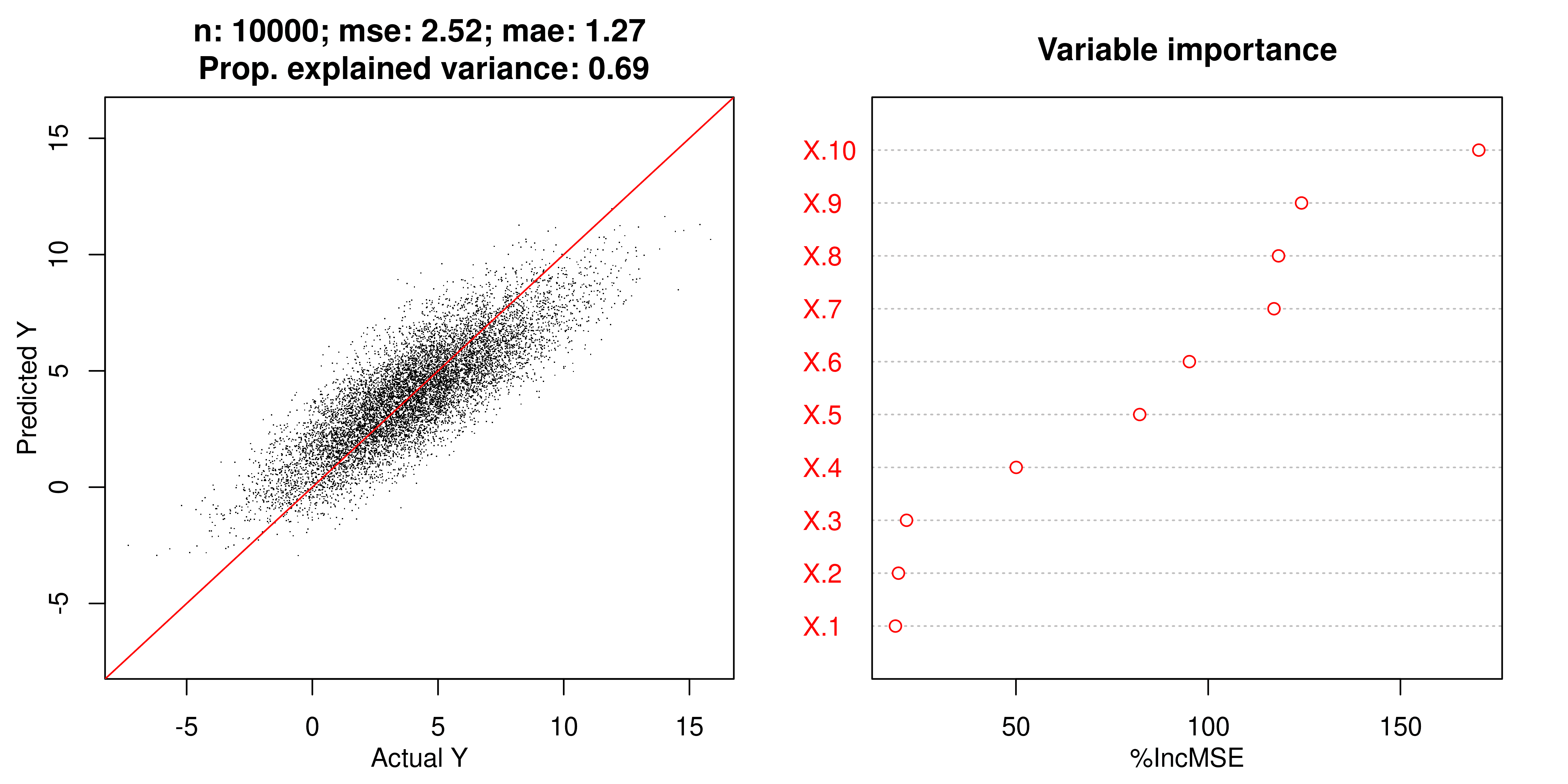}
\includegraphics[clip,trim=0 0 0 0,angle=0,width=1\textwidth]{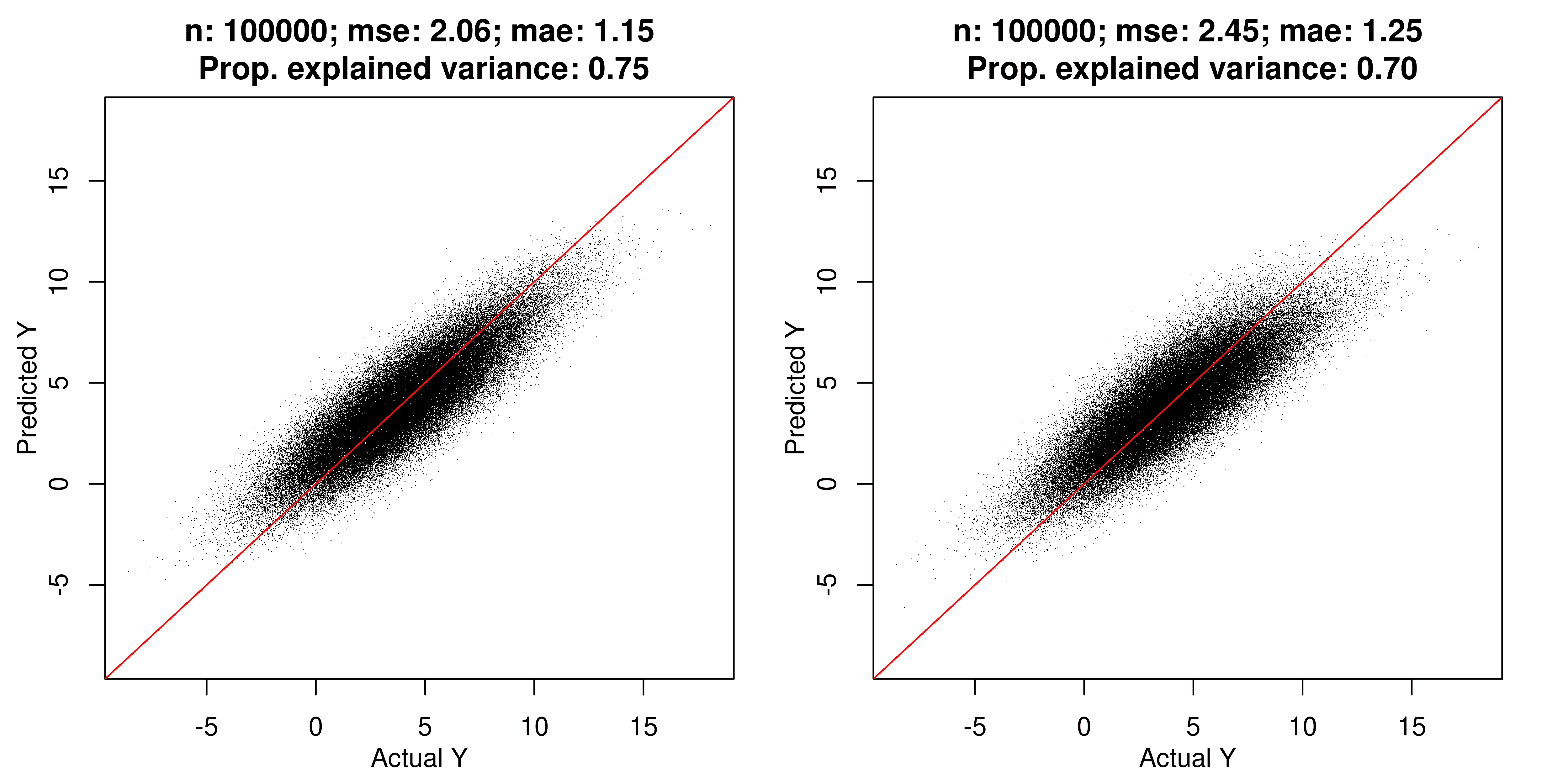}
\caption{
A summary of the results obtained with Breiman's random forest and with the two-armed random forest. {\it Top panel}: scatter-plot, statistics and variable importance plot summarizing the performance of the random forest based on a data set of size 10,000. {\it Lower panel}: scatter-plots and statistics summarizing the performance of the two-armed random forest (left) and of the random forest based on a data set of size 100,000.}
\end{center}
\end{figure}

\begin{figure}[hbtp]\label{figure2}
\begin{center}
\includegraphics[clip,trim=0 0 0 0,angle=0,width=1\textwidth]{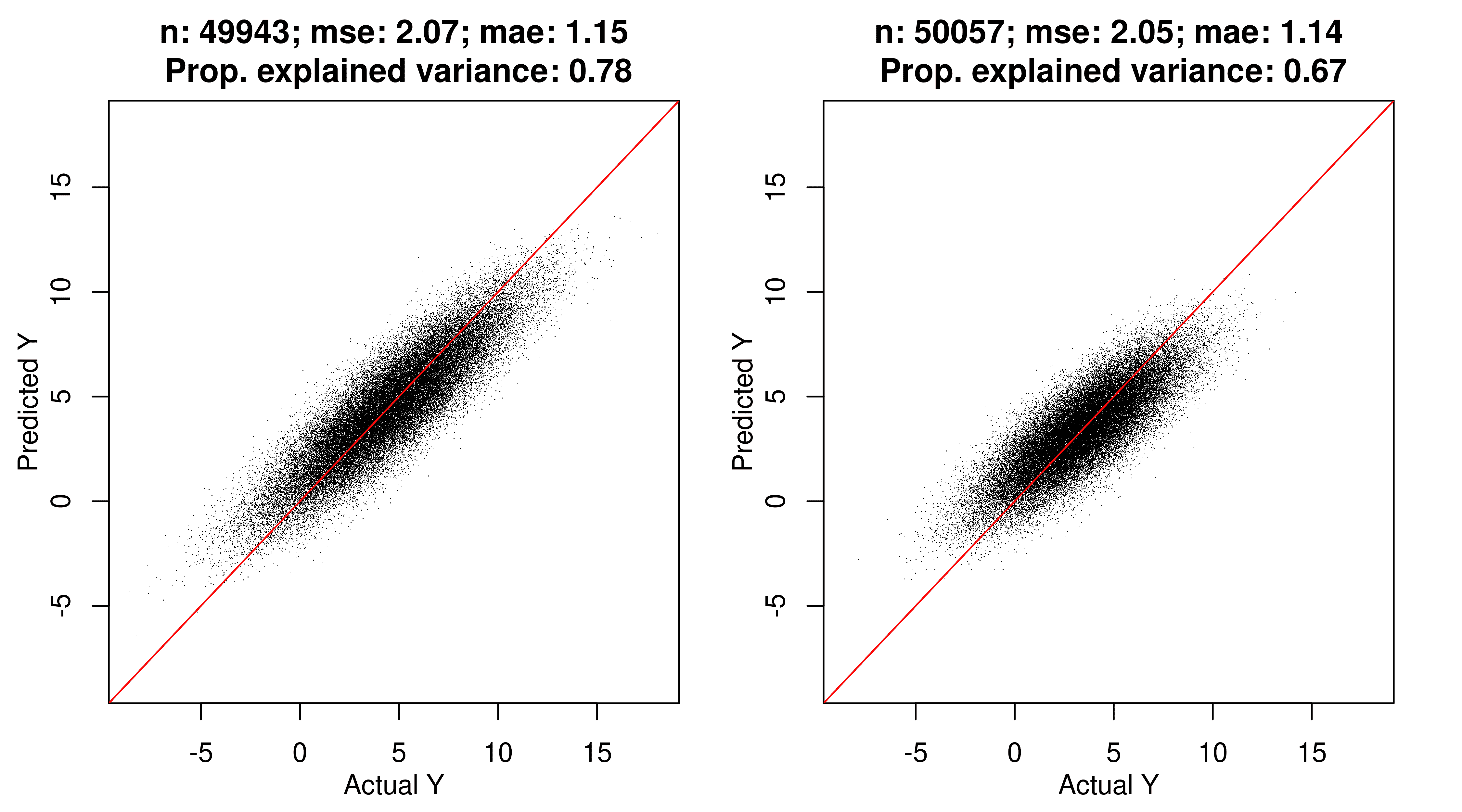}
\includegraphics[clip,trim=0 0 0 0,angle=0,width=1\textwidth]{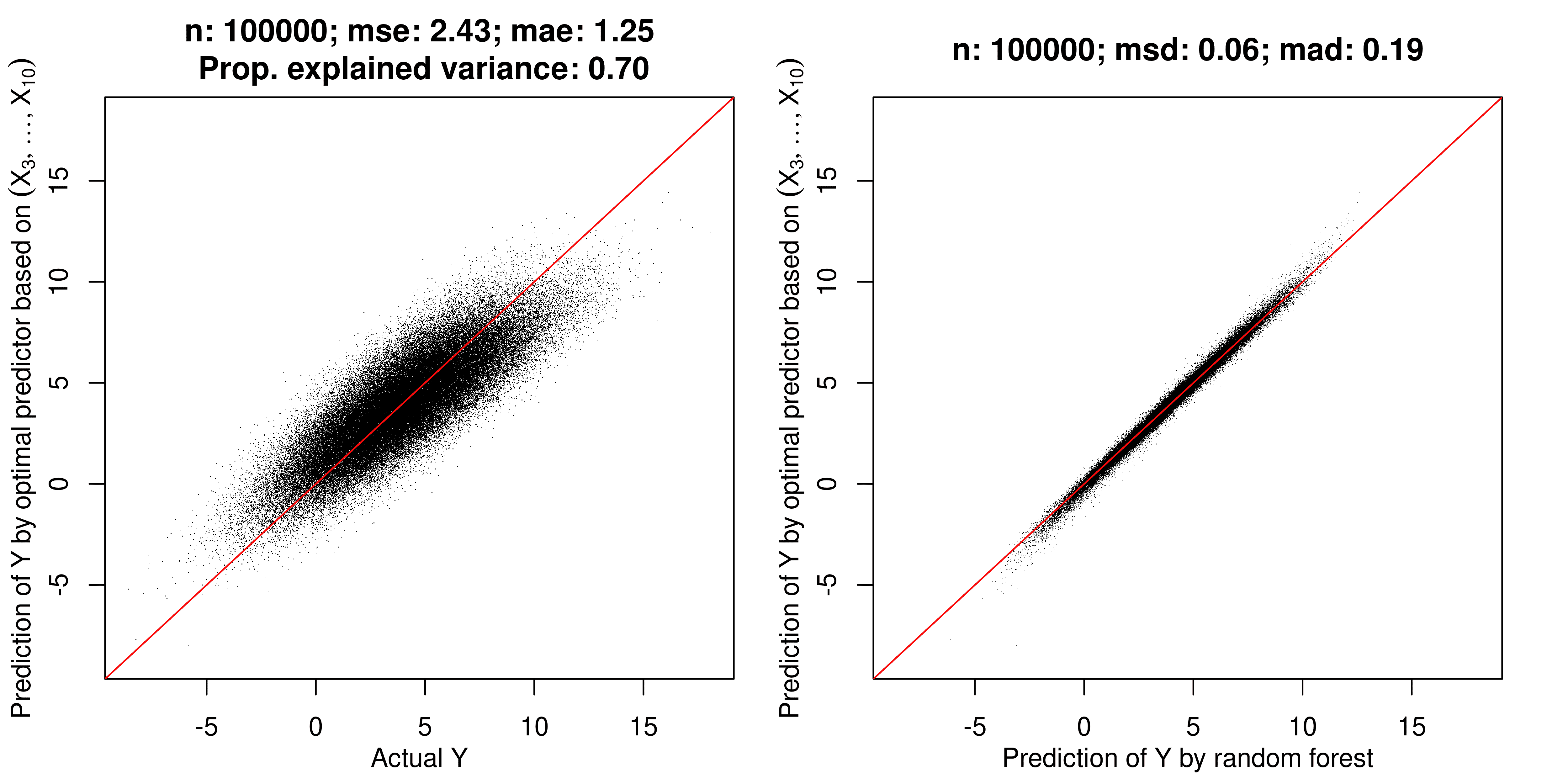}
\caption{
{\it First panel}: scatter-plots and statistics summarizing the performance of the two random forests that make up the two-armed random forest based on a training data set of size 100,000. {\it Second panel}: left, scatter-plot and statistics summarizing the performance of the optimal predictor of $Y$ based on $(X_3,\ldots,X_{10})$, which is expected to perform similarly to the random forest predictor; right, scatter-plot and statistics (mean square difference, mean absolute difference) comparing the predictions of the random forest with those of the optimal predictor of $Y$ based on $(X_3,\ldots,X_{10})$.}
\end{center}
\end{figure}

\indent
Table 1 shows some estimates of the mean-square error of random forests with the default value of {\tt mtry} and with ${\tt mtry}=1$ and of the two-armed random forest.
Although at two decimal places the two-armed random forest does not seem to get closer to the theoretical optimum with 
the larger values of $n$, that must be due to the number of trees not being increased beyond 1000 (consistency requiring of course an infinite number of trees). On the other hand, with the larger values of $n$ the performance of the random forests remains almost uniformly away from the optimum. Interestingly,
with ${\tt mtry}=1$ the random forest ranks $X_1$ and $X_2$ as the most important variables already when $n\ge 100,000$ (when $n=10,000$ they come very close to each other in third and sixth places), but its performance is worse than with the default value of ${\tt mtry}=3$. With the latter, $X_1$ and $X_2$ 
appear slightly above $X_3$ when $n\ge 200,000$. As shown later, the correct estimation of the importance of $X_1$ and $X_2$ places them between  $X_7$ and $X_8$, not at the top as the choice of ${\tt mtry}=1$ suggests. In any case, even if the importance of `hidden predictors' such as $X_1$ and $X_2$ is not always  detectable with the standard choice of {\tt mtry} (depending on the sample size and on the relative strength of the predictors) it is clear that they have good chances of being detected with ${\tt mtry}=1$, unless perhaps the number of trees is small compared to the number of variables. 

\vspace{-0.1cm}
\begin{table}
\begin{center}
\caption{\small Estimates based on test sets of size $n$ of the mean-square error of two versions of random forest (RF) and of the two-armed random forest constructed with training sets of size $n$.}
\begin{tabular}{c|cccc}
\hline
$n$       & RF with ${\tt mtry}=3$ & RF with ${\tt mtry}=1$ & Two-armed RF  \\ \hline
10,000   & 2.52 & 2.81 & 2.13 & \\
100,000 & 2.45 & 2.79 & 2.06 & \\
200,000 & 2.45 & 2.75 & 2.06 & \\
500,000 & 2.45 & 2.75 & 2.06 & \\
\hline
\end{tabular}
\end{center}
\vspace{-0.5cm}
\end{table}

\subsection{Statistics of variable importance and variable usage}
The gap between the mean-square error of the random forest and that of the optimal predictor can be very large provided the two branches of model (\ref{equation8}) are very different. For example, when
${\boldsymbol\beta}=-{\boldsymbol\alpha}$ and ${\boldsymbol\alpha}$  is as above, random forests based on samples of sizes as large as 100,000 and 200,000 have mean-square errors of about 20, while the corresponding two-armed random forest has a mean-square error of about 2.07. In this case the importance of each of $X_1$ and $X_2$ is about two to three times as great as that of the other variables, but this does not mean that $X_1$ and $X_2$ are used a lot more in the trees when
${\boldsymbol\beta}=-{\boldsymbol\alpha}$ than when ${\boldsymbol\beta}=3{\boldsymbol\alpha}/4$ because variable importance does not account for the number of times a variable has been used to define a terminal node; rather, it is more a reflection of the fact that distinguishing the two branches of model (\ref{equation8}) is so necessary that whenever splits on $X_1$ and $X_2$ occur they dramatically improve the accuracy within the descendent terminal nodes. In fact, we observe that the probability of there occurring a split on $X_1$ or $X_2$ is about 0.05 during most of the construction of a tree under both versions of the model (it is much lower at the beginning of the construction, especially when ${\boldsymbol\beta}=3{\boldsymbol\alpha}/4$); this is clear from the first panel of figure 3, which shows the proportion of splits on $X_1$ or $X_2$ as a function of the order of the node split during the construction of the trees.\\
\begin{figure}[hbtp]\label{figure3}
\begin{center}
\includegraphics[clip,trim=5 0 0 7.5,angle=0,width=0.475\textwidth]{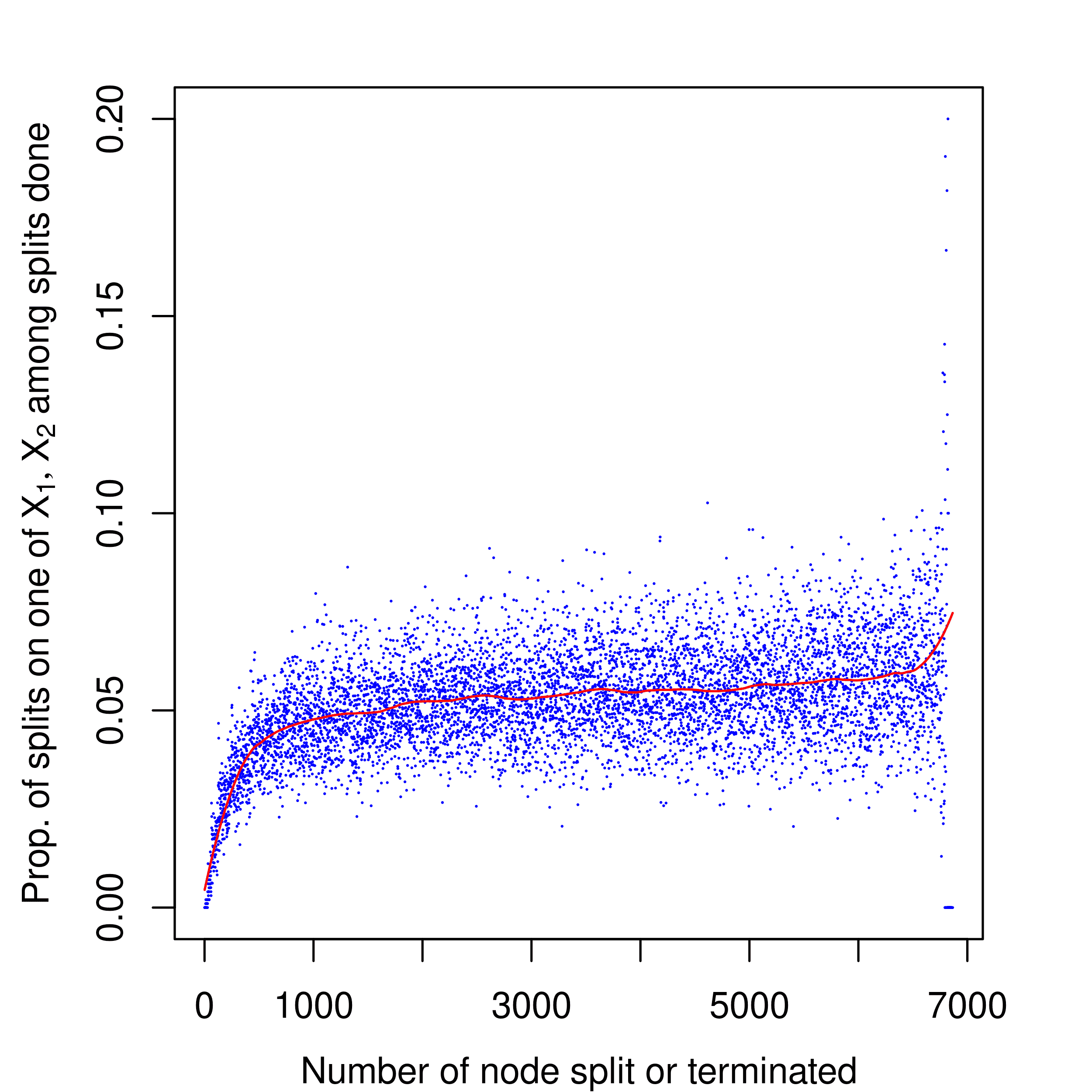}
\includegraphics[clip,trim=5 0 0 7.5,angle=0,width=0.475\textwidth]{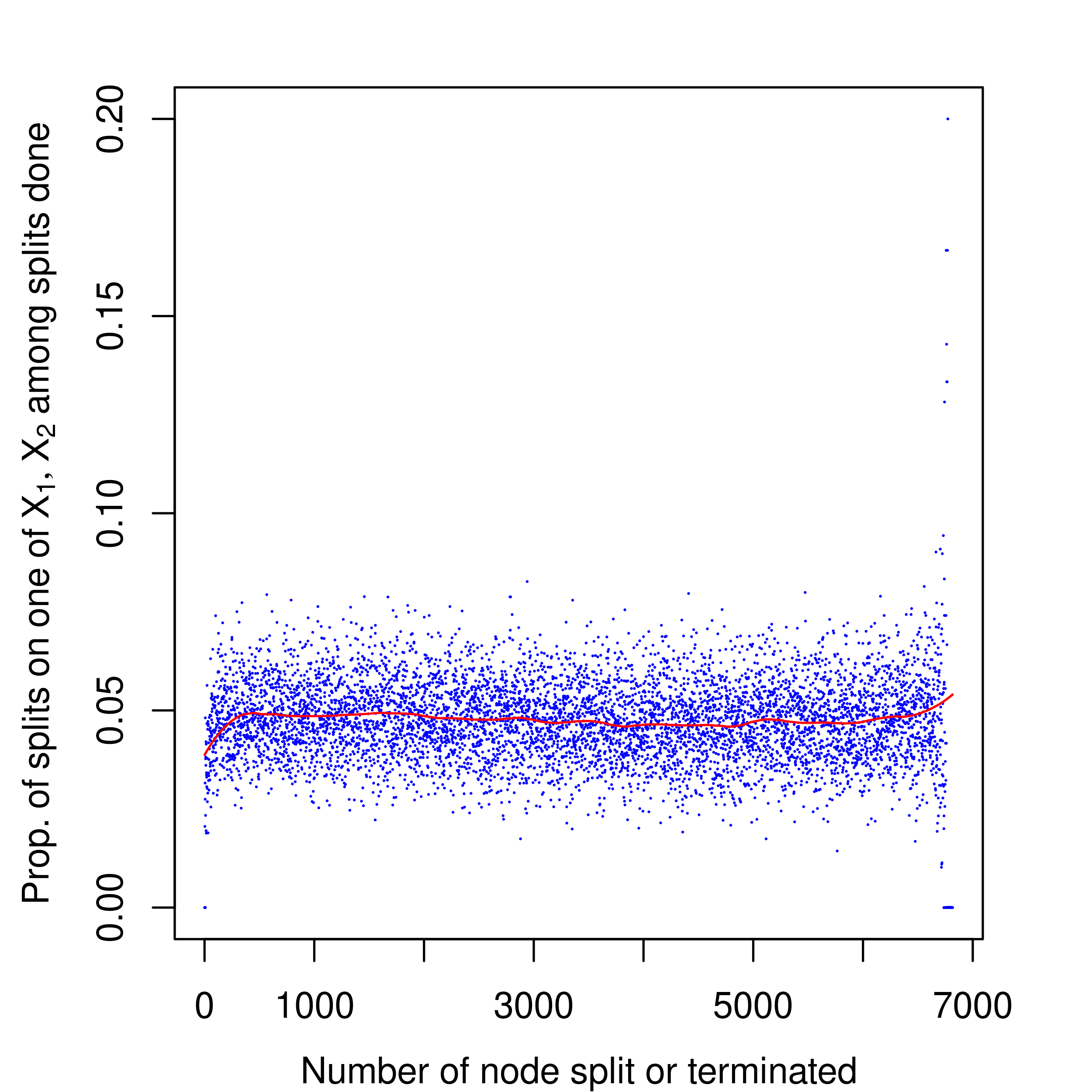}\\
\includegraphics[clip,trim=5 0 0 7.5,angle=0,width=0.475\textwidth]{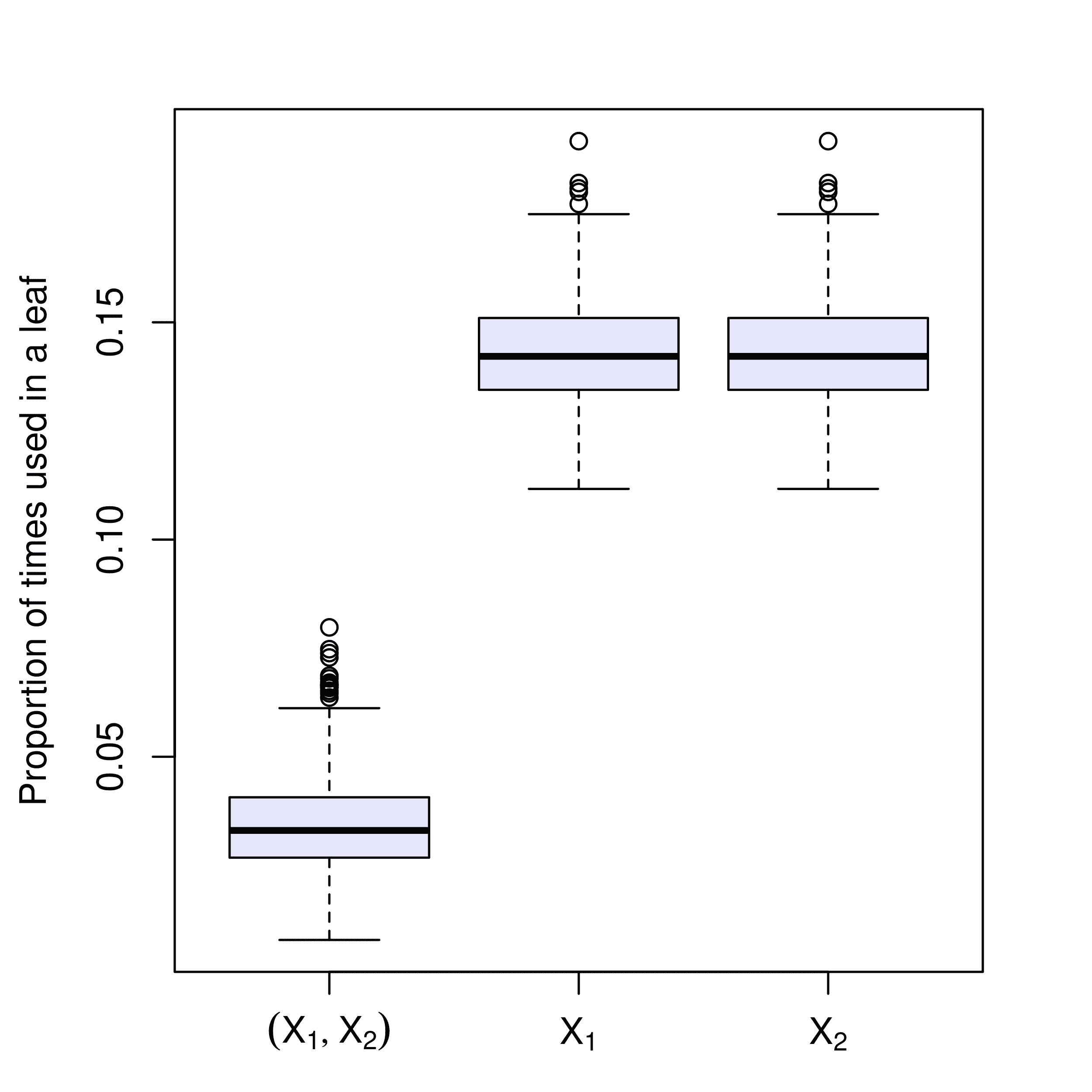}
\includegraphics[clip,trim=5 0 0 7.5,angle=0,width=0.475\textwidth]{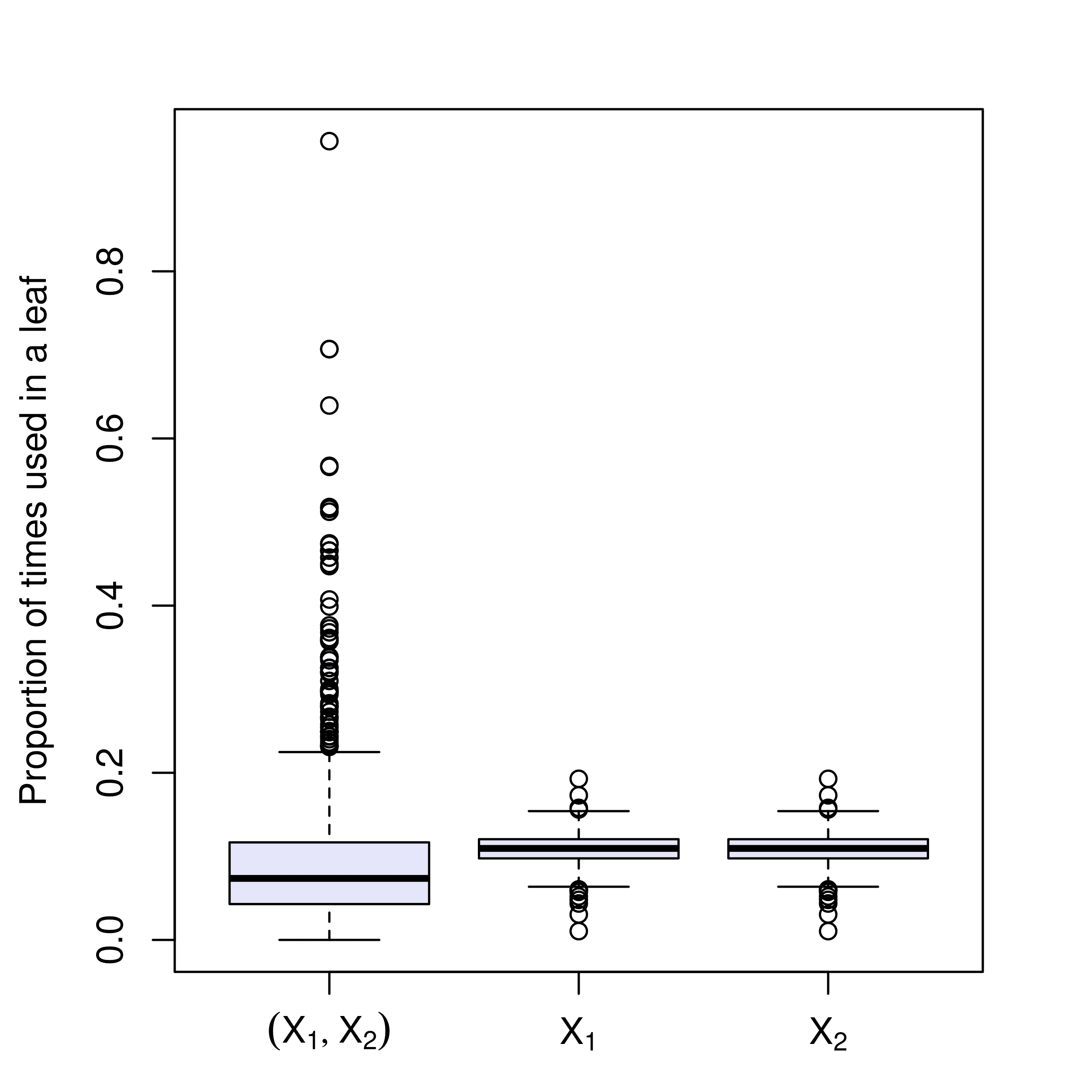}
\caption{
{\it First panel}: Proportion of splits on $X_1$ or $X_2$ as a function of the nodes already split or terminated in 1000 trees grown from a simulated sample of size 10,000 from model (\ref{equation8}) with ${\boldsymbol\beta}=3{\boldsymbol\alpha}/4$ (left) and ${\boldsymbol\beta}=-{\boldsymbol\alpha}$: As trees are built, new nodes are appended to them (or else existing nodes are terminated because they cannot be split any further); these operations, labelled from one to the number of nodes in the largest tree, are represented by the horizontal axis. At the time of the $k$-th operation one calculates the proportion of trees split on $X_1$ or on $X_2$ among all trees split at that operation, and this is represented on the vertical axis.
{\it Second panel}: Proportion of times that $X_1$ and $X_2$ together or on their own are used in the terminal nodes (`leaves') of 1000 trees constructed from bootstrap samples of a simulated sample of size 10,000 from model (\ref{equation8}) with ${\boldsymbol\beta}=3{\boldsymbol\alpha}/4$ (left) and ${\boldsymbol\beta}=-{\boldsymbol\alpha}$. The trees are constructed as in the random forests (nodes being split as long as they contain five or more distinct observations) except that all the 10 predictor variables are tried for a split at each node (while in the forests only three randomly chosen variables are tried).}
\end{center}
\end{figure}
\indent
It would be interesting to check how often $X_1$ and $X_2$, together or on their own, are involved in the terminal nodes of a tree, that is, how many cells of the associated partition represent a restriction on the domain of $X_1$, $X_2$ or $(X_1,X_2)$. We are unable to obtain such information from the output of the {\tt randomForest} function of \cite{LiawWiener:2002}, but we can use the {\tt R} package {\tt rpart} of \cite{TherneauAtkinson:2019} to compute the number of terminal nodes involving $X_1$ and $X_2$ together and on their own in a tree grown as the trees in a random forest except that all the 10 predictor variables (and not just a random subset of three) are tried for a split at each node; since the performance of the random forest in our examples remains practically the same if the 10 variables are tried at each split, this should yield reliable information about how often trees in a forest with the default value of {\tt mtry} use $X_1$ and $X_2$. The box plots of figure 3 summarize the distributions of the proportion of terminal nodes in trees involving $X_1$ and $X_2$, on their own and together, obtained from 1000 trees constructed with bootstrap samples drawn from samples of size 10,000 simulated according to the two versions of model (\ref{equation8}) we are considering. Typically, fewer than $15\%$ of the terminal nodes in a tree use one of the two variables, and fewer than $10\%$ use both; the only essential difference between the two models is that when ${\boldsymbol\beta}=-{\boldsymbol\alpha}$ there are a few more trees with a higher percentage of terminal nodes involving both $X_1$ and $X_2$. It may also be interesting to look at the numbers of observations contained in the terminal nodes that involve both $X_1$ and $X_2$: on average, fewer than $4\%$ of the data in the first case
(${\boldsymbol\beta}=3{\boldsymbol\alpha}/4$) and fewer than $7\%$ in the second take advantage of the values on $(X_1,X_2)$; in contrast, the two-armed random forest predictor, which appears to approach the minimum mean-square error, uses the values on $(X_1,X_2)$ to predict every response.\\
\indent
Together with the arguments given in section 2, the observations made in this section suggest that monitoring variable importance in conjunction with the proportion of terminal nodes involving each variable in a tree is a means of diagnosing the presence of variables such as $X_1$ and $X_2$. Such a procedure certainly works in the case ${\boldsymbol\beta}=-{\boldsymbol\alpha}$ of the present model because of the great discrepancy between the frequency with which $X_1$ and $X_2$ are used and their importance; in a case like ${\boldsymbol\beta}=3{\boldsymbol\alpha}/4$ the measure of variable importance provided by random forest may be misleading (depending on the sample size) and hide the potentiality of $X_1$ and $X_2$ with the default value of {\tt mtry}, but not with ${\tt mtry}=1$.\\

\subsection{The variable importance of $X_1$ and $X_2$}
We finish this analysis by determining variable importance in the two-armed random forest more correctly.
In order to do this we note in the first place that the importance of $X_j$ in a predictor $\Pi$ based on $(X_1,\ldots,X_d)$ may, for example, be defined as the number
\[
I_j(\Pi)=100\times\frac{\mathbb{E}\left[L\left(Y,\Pi(X_1,\ldots,\tilde X_j,\ldots,X_d)\right)\right]-{\bf e}(\Pi)}
{{\bf e}(\Pi)},
\]
%\[
%I_j(\Pi)=
%100\times\frac{\mathbb{E}\left[L\!\left(Y,\Pi(X_1,\ldots,\tilde X_j,\ldots,X_d)\right)\right]-
%\mathbb{E}\left[L\!\left(Y,\Pi(X_1,\ldots,X_j,\ldots,X_d)\right)\right]}
%{\mathbb{E}\left[L\!\left(Y,\Pi(X_1,\ldots,X_j,\ldots,X_d)\right)\right]},
%\]
where ${\bf e}(\Pi):=\mathbb{E}\left[L\!\left(Y,\Pi(X_1,\ldots,X_d)\right)\right]$,
$L$ is a loss function (typically $L(s,t)=(s-t)^2$), and $\tilde X_j$ has the same distribution as $X_j$ but is independent of the other $X_i$s (if $\Pi$ depends on a training data set the expectations are conditional on that set).\footnote{This is just one of the possible definitions of the importance of a variable; for a recent overview of other definitions and methods of estimating variable importance see \cite{LohZhou:2021}.}
Secondly, $I_j (\Pi)$ can be estimated from data, say by estimating ${\bf e}(\Pi)$ with a large test set and averaging many (say 1000) estimates of $\mathbb{E}[L(Y,\Pi(X_1,\ldots,\tilde X_j,\ldots,X_d))]$ computed with perturbed versions of the test set obtained by randomly permuting in it the data on the $j$-th predictor variable.\footnote{This is not the same as the method used in Breiman's random forest, which for economy computes such estimates per tree and then averages them, but in our experience the two methods generally provide a similar ranking of importance.} Finally, note that $I_j(\Pi)$ depends on $\Pi$ and not just on
$(X_1,\ldots,X_d)$, so a suboptimal $\Pi$ need not make the best possible use of the predictor variables and may unduly deflate or inflate their importance. The {\it optimal predictor}, on the other hand, makes the best possible use of every variable and therefore $I_j (\Pi)$ with $\Pi$ as the optimal predictor provides a more faithful measure of variable importance, and so does an estimator of it obtained from a {\it consistent predictor} when the estimator and the consistent predictor are based on large samples.\\
\indent
Of course, in real-life problems one ignores the optimal predictor and is often unsure about the consistency of the predictor in hand, so one can seldom be completely sure of estimating the {\it correct variable importance}, which here we take to be represented by $I_j (\Pi)$ with $\Pi$ as the optimal predictor.
But in our case we know that the optimal predictor under model (\ref{equation8}) is
$\Pi^*(X_1,\ldots,X_d)=\delta_{X_1,X_2} (\alpha_3 X_3+\cdots+\alpha_d X_d)+\{1-\delta_{X_1,X_2}\}(\beta_3 X_3+\cdots+\beta_d X_d)$. Since the two-armed random forest is so close to the optimal predictor, we trust the variable importance estimated from it much more than we do the variable importance estimated from a random forest. Figure 4 shows estimates of the $I_j (\Pi^*)$s, computed as indicated in the first paragraph of this subsection (so that variable importance really represents a percent worsening of the predictor's accuracy that results from random permutations of the data on a variable),
from two-armed random forests constructed with training data sets of size 10,000 simulated according to (\ref{equation8}) in the two cases we have been considering. While in the case ${\boldsymbol\beta}=-{\boldsymbol\alpha}$ the essential aspect of the ranking coincides with the one proposed by random forest (except that $X_1$ and $X_2$ are now recognized as even more important than the rest), in the case ${\boldsymbol\beta}=3{\boldsymbol\alpha}/4$ the new ranking shows that $X_1$ and $X_2$ are much more important than the random forest had suggested with the default value of {\tt mtry} (cf. the top right plot of figure 1) and less important than it had suggested with
${\tt mtry}=1$. Accordingly, the benefit of recognizing the importance of $X_1$ and $X_2$ is not that great in the latter case and enormous in the former. 

%\vspace{-0.85cm}
\begin{figure}[hbtp]\label{figure4}
\begin{center}
\includegraphics[clip,trim=0 0 0 0,angle=0,width=0.45\textwidth]{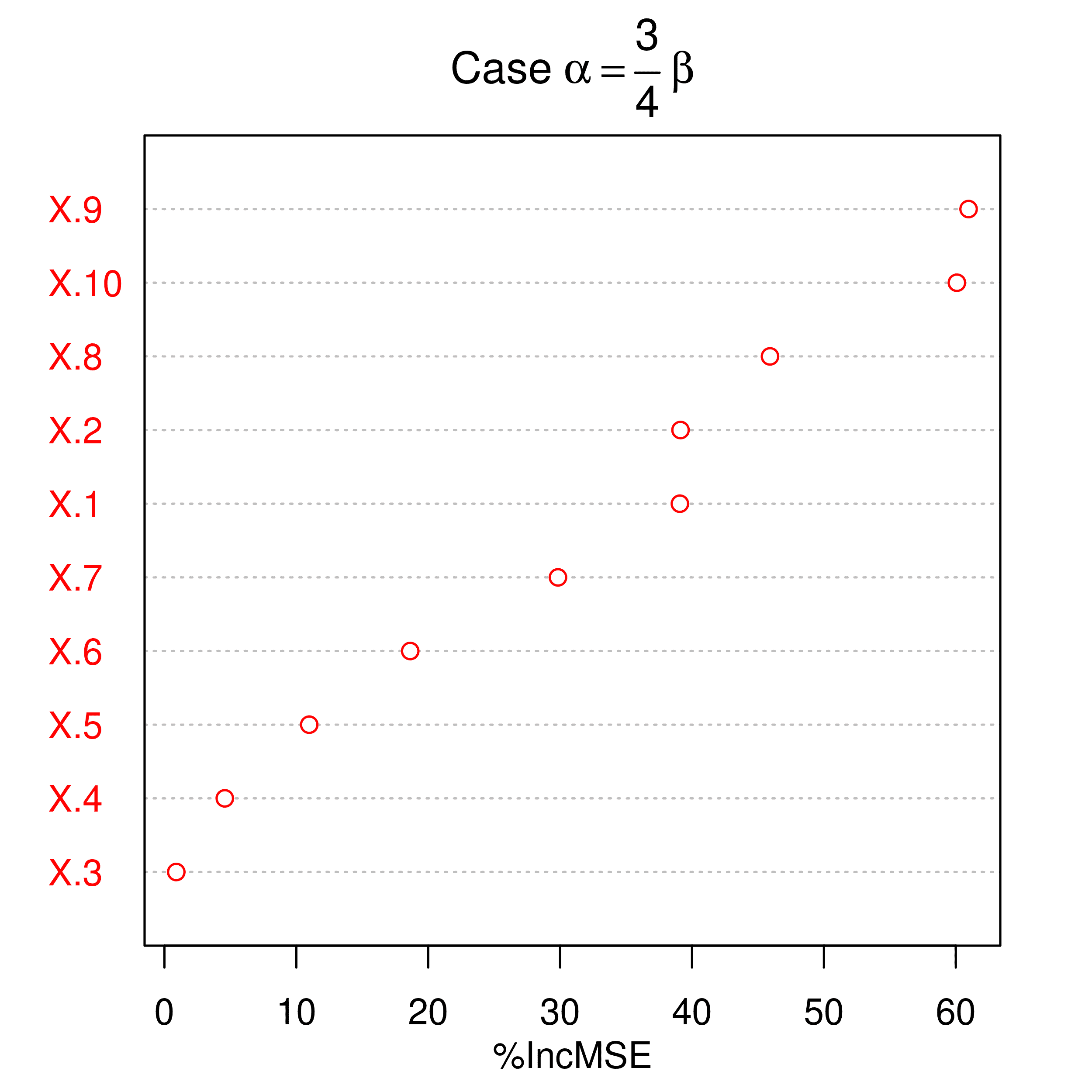}
\includegraphics[clip,trim=0 0 0 0,angle=0,width=0.45\textwidth]{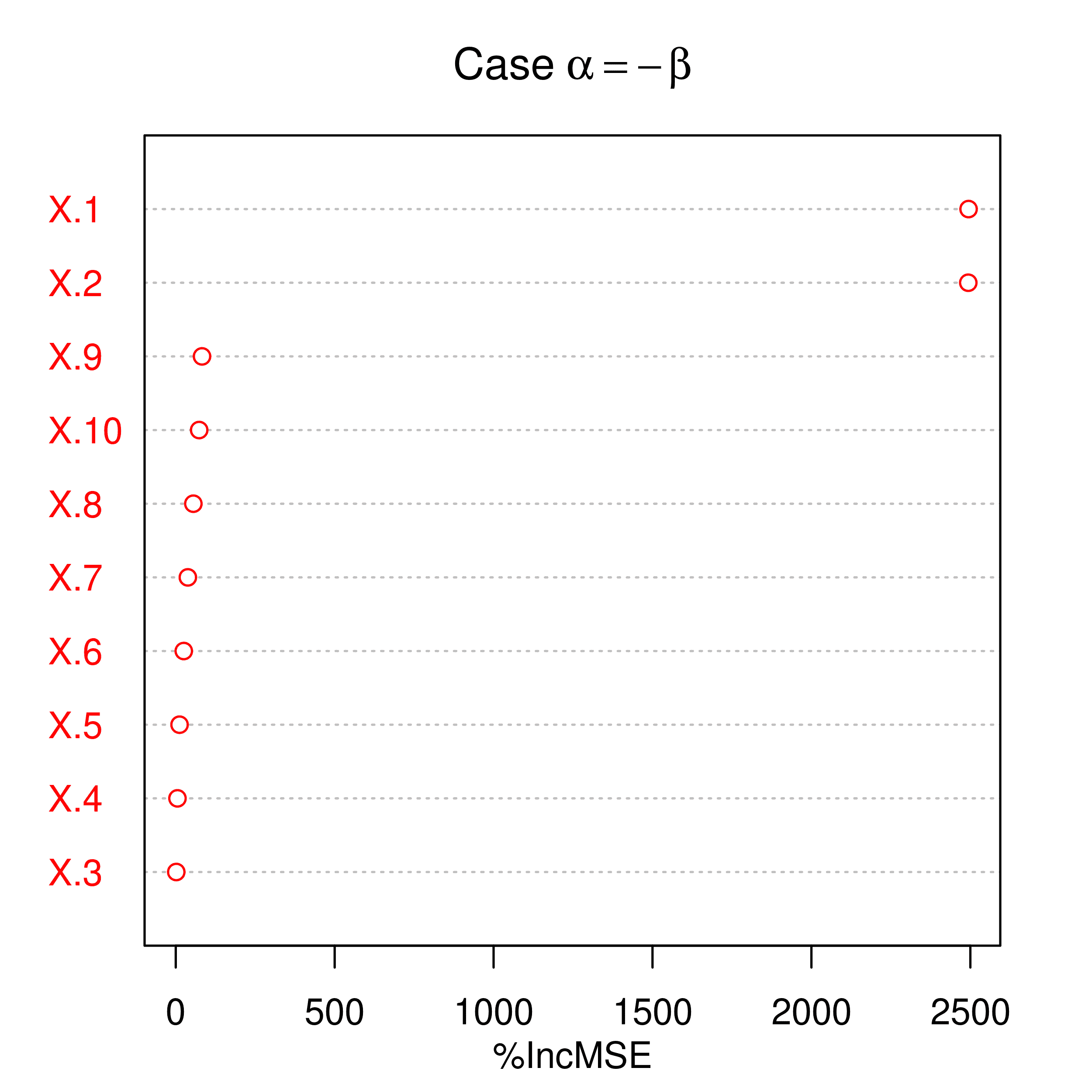}
\caption{
Variable importance according to the two-armed random forest constructed with a sample of size 10,000. Importance is estimated as the average percent increases in the mean-square error of the predictions on a test set of size 10,000 that result from perturbing the same test set by 1000 random permutations of the data on each predictor variable at a time. The ranking of the variables on the left should be compared with that shown in figure 1.}
\end{center}
\end{figure}

\section{Discussion}

We have seen that there are data-generating models which neutralize a powerful strategy of Breiman's random forest, namely the one by one identification of the more predictive variables. Although it is unlikely that real data follow such models closely, examples from genetics in which haplotypes rather than genotypes predict phenotypes suggest that some data may follow them approximately. We have seen that `many-armed' random forests can be much better than random forests at predicting responses from such models, so it is not implausible that many-armed random forests like the one used in section 3 will perform better than random forests on some real data sets. Although our exposition has concentrated on the prediction of a numerical variable, these observations apply as well to the prediction of nominal variables (i.e. to `classification problems').\\
\indent
The general approach to constructing a many-armed random forest consists of identifying `hidden predictor variables', such as the $X_1$ and $X_2$ in the first example of section 2, or more generally the coordinates of ${\mathbf X}_1$ and ${\mathbf X}_2$ in the last example, creating an initial partition on those variables, and constructing a random forest within each member of that partition. As a means of identifying possible hidden predictors we propose looking at the usual measures of variable importance in conjunction with measures of {\it variable usage} which keep track of how frequently variables are used in trees, the motivation for this being that a strong predictor variable that `needs help' from other variables in order to be included in trees may appear as important and yet be used very little, without this becoming visible from variable importance alone. Although its success depends on the relative strength of the hidden predictors, on the number of variables, on the sample size and on many other things, in applications there is certainly no harm in monitoring variable usage in addition to variable importance.
There must be ways of combining the two types of measure other than those we have used in our illustrations of section 3, and better ways should become apparent if {\it statistics of variable usage} be made available in implementations of random forest; as far as we know, current implementations do not provide `ready-made' statistics of variable usage, but we think that they could easily do so without substantial changes in their code and functionality.\\ 
\indent
In our illustrations of section 3 the hidden predictors are binary and the initial four-part partition on them is essentially given. When the hidden variables are many and/or take a large number of values the procedure is necessarily more complicated and there are various ways of specifying the initial partition, not all of them being equally feasible nor leading to equally accurate results. For example, one might think of creating partitions of `$k$-cells' whenever there are $k$ potential hidden predictors and choosing the one yielding more accurate predictions on a training data set (as in the variant of tree and forest predictors proposed in section 20.14 of \cite{DevroyeGyorfiLugosi:1996} and in section 6 of \cite{BiauDevroyeLugosi:2008}, already mentioned in our introduction). Or one might take the much simpler and faster approach of splitting each of the potential hidden predictors at their medians or quartiles and using the resulting partition of $k$-cells.
Finding ways of performing feasible and favourable initial partitions constitutes a research project in itself and has not even been attempted here.
As far as we know, the implementations of random forest currently available do not permit the creation of
initial partitions based on a list of variables provided by the user; our results suggest that possibilities for doing this may sometimes be useful.

\acks{The author is indebted to several reviewers and editors for comments that helped improving the paper.}

\vskip 0.2in
\bibliography{ConvergenceOfRandomForest}

\end{document}